# Evaluating the Accuracy of Chatbots in Financial Literature

Orhan Erdem[a], Kristi Hassett[b], Feyzullah Egriboyun[c]

[a]: (Corresponding Author) University of North Texas, Advanced Data Analytics Department, 1155 Union Circle #310830 Denton, TX 76203-5017, ORCID: 0000-0002-0173-2364, email: orhan.erdem@unt.edu

[b]: University of North Texas, MS in Advanced Data Analytics Candidate, Advanced Data Analytics Department, 1155 Union Circle #310830 Denton, TX 76203-5017, email: KristinHassett@my.unt.edu

[c]: Hult International Business School, Hult House East, 35 Commercial Rd, London E1 1LD, email: Feyzullah.egriboyun@faculty.hult.edu

# Assessing the Accuracy of Chatbots in Financial Literature

## Abstract

We evaluate the reliability of two chatbots, ChatGPT (4o and o1-preview versions), and Gemini Advanced, in providing references on financial literature and employing novel methodologies. Alongside the conventional binary approach commonly used in the literature, we developed a nonbinary approach and a recency measure to assess how hallucination rates vary with how recent a topic is. After analyzing 150 citations, ChatGPT-4o had a hallucination rate of 20.0% (95% CI, 13.6%-26.4%), while the o1-preview had a hallucination rate of 21.3% (95% CI, 14.8%-27.9%). In contrast, Gemini Advanced exhibited higher hallucination rates: 76.7% (95% CI, 69.9%-83.4%). While hallucination rates increased for more recent topics, this trend was not statistically significant for Gemini Advanced. These findings emphasize the importance of verifying chatbot-provided references, particularly in rapidly evolving fields.

## 1. Introduction

*"Mine eyes are made the fools o' the other senses."*

Macbeth

Artificial intelligence (AI) was developed to mimic human cognitive processes, which was different from traditional computer software designs. This type of computer programming involves a new set of challenges, including the phenomenon of "hallucinations" created by AI. These hallucinations refer to outputs from large language models (LLMs), a specialized subset of AI, that are not based on factual data. Although (Maleki et al., 2024) reviewed fourteen databases and found no consensus on the precise definition or characterization of AI hallucination, researchers have documented various instances of AI chatbots generating nonfactual data.

(Chen & Chen, 2023) studied hallucinations in healthcare and classified the journal as fake if any of the components, title, authors, publication year, volume, issues, or pages were not accurate. They found that GPT 3.5 had a 98.1% hallucination rate and that GPT-4 had a 20.6% hallucination rate. They showed that broader topics had fewer hallucinations than narrow topics. Other papers took a different approach to differentiating between real vs hallucination. (Chelli et al., 2024) used databases that contained randomized trials for rotator cuff pathology published between 2020 and 2021. They found hallucination rates of 39.6% for GPT-3.5, 28.6% for GPT-4 and 91.4% for Gemini[1]. While they researched medical papers in two specific years, we are researching finance topics without a date restriction, our results are similar. (Magesh et al., 2024) labeled responses as "correctness" or "groundedness." Correctness was determined if the response was both real and relevant. Groundedness was determined by the quality of the output according to legal case law. The authors included a degree of hallucination within these types to include partial hallucinations. The hallucination rates varied with both Lexis+ AI and Ask Practical Law AI at 17%, Westlaw at 33% and ChatGPT at the highest hallucination rate of 43%.

LLMs are often able to create output faster than humans. (Nazzal et al., 2024) selected two medical topics about the skeleton and had three groups of authors writing on each topic for a total of eight papers. The author groups included human, AI, and AI with citations from human papers. They found that while AI reduces writing time, it often lacks accuracy and may lead to plagiarism. The findings underscore that AI can expedite the initial drafting process but requires substantial human editing to ensure accuracy and originality. Unlike in our study, (Gao et al., 2023) had ChatGPT create fake scientific abstracts, and human reviewers categorize the abstracts as either real or fake. They identified the abstracts created by AI 68% of the time. They also incorrectly labeled the AI-generated abstracts as real 14% of the time. (MIT Sloan, 2023) succinctly identified three reasons why hallucinations could occur, including the need for additional training data, the idea that AI was designed to predict and find patterns, and the fact that it was not designed to distinguish between real and fake information. These reasons for hallucinations are not straightforward solutions for reducing hallucination rates.

A possible solution was proposed by (Xu et al., 2023) to reduce hallucinations in language translation. They studied automated translation from one language to the next in a "neutral" way.

---

[1] Google rebranded Bard to Gemini on March 21, 2023

They proposed a solution to reducing hallucinations by intentionally prompting with a small inconsistency in the text. They found that identifying a hallucination was more effective than comparing the text to the generated output. This method of tricking AI into correcting the prompt might not work for all industries but is worthy of investigating. (Nature Medicine Editorial, 2023) stated that the use of LLMs can assist in the medical field, but caution should be taken when relying on AI. Like in our study in the finance literature, they suggest relying on LLMs to perform diagnoses without the use of human expertise and logic. There have been many studies on hallucinations, each using a different definition, technique and approach. Our study evaluated the pieces of each response to determine the degree of hallucination as well as other anomalies we found, such as the year of the article.

In a study that closely resembles our study, (Walters & Wilder, 2023) prompted ChatGPT-3.5 and ChatGPT-4 to create papers with 2000 words and five citations per paper. The prompts used 42 topics with a broad range of subjects. Their purpose was to identify "factually incorrect responses (hallucinations)" and errors in the APA citations. Their results showed that GPT-3.5 had a higher rate of 55% hallucination, while GPT-4 had 18%. Among the non-hallucinated responses GPT-3.5 had a higher error rate in the citation of 43% and GPT-4 had 24%. While their prompts could be considered more detailed by asking the chatbot to "act as" a researcher, write a short academic paper and provide citations, their hallucination rate for GPT-4, 18%, is similar to ours, 20%. Potential differences in hallucination rates could be related to age and type of topic. In this study, we mainly focus on the financial literature (with a wide range of dates) rather than a broader array of topics. We also explored the effect of the recency of the topic on the hallucination rate.

Very few papers in the economics and finance literature have discussed hallucinations. In one of these papers, (Buchanan et al., 2024) created a similar study to this paper by using ChatGPT-3.5 and GPT-4 to query economic topics from the Journal of Economic Literature. They used three prompts, each with different levels of detail. The authors determined that the hallucination rate for ChatGPT-3.5 exceeded 30%, while ChatGPT-4 generated incorrect citations at a rate higher than 20%. Their results correspond to our scale mentioned below. Our topics are focused on financial literature than on general economics (for example, "economic theory" as a general theme versus "incentives"). The authors added that hallucinations grew in both GPT-3.5 and GPT-4 when the prompts were more specific. This result contrasts with the results of Chen et al., where broader topics had fewer hallucinations than narrow topics.

Our paper distinguishes itself from the aforementioned studies in several keyways. First, we provide a comparative analysis of the outputs from three of the most recent chatbot models: ChatGPT-4o, o1-preview, and Gemini Advanced (Gemini A., hereafter). ChatGPT-4o was released on May 13, 2024, Gemini A. was released on May 14, 2024, and o1-preview on September 14, 2024. Consequently, this study is among the first, if not the earliest, to document the performance of the o1-preview model. According to Open AI o1-preview, which performs better than ChatGPT-4o in many tasks, "On one of our hardest jailbreaking tests, GPT-4o scored 22 (on a scale of 0-100), while our o1-preview model scored 84." Moreover, while prior research predominantly investigates medical literature, our focus lies on finance literature. Additionally, we introduced two novel metrics to measure hallucination. The first extends beyond a simple binary scale by incorporating degrees of hallucination, whereas the second, a 'recency measure', quantifies the variance in hallucination depending on the "age" of the topic.

The paper is organized as follows: Section 2 discusses the collection of data, Section 3 explains the methodology, Section 4 documents the results and Section 5 concludes.

## 2. Data

In this study, we tasked three of the most recent chatbots, ChatGPT-4o, o1-preview, and Gemini A., by providing ten unique references for fifteen topics in finance.

After reviewing the financial literature classification via JEL codes, we determined that these fifteen topics span the bulk of finance literature. Thus, the chatbots each listed a total of 150 references. The following is a prompt we gave for the first topic (see Table 1):

*"Please provide a literature review of 10 unique references from journal articles that investigate the effects of financial incentives on decision-making."*

For the remaining fourteen topics, we modified the underlined segment of the prompt accordingly. We then utilized multiple databases and other web searches to verify the authenticity of the listed papers. The data were collected from August 1, 2024, to September 14, 2024.

# 3. Methodology

### 3.1 Hallucination metrics

We verified each cited journal article by confirming its existence in the referenced journal and by searching its title using Google Scholar, and publisher websites. We checked the article's title, journal name, authors, and publication year. We labeled an entry "hallucination" if any of these four features did not match reality. To quantify the degree of hallucination, we used two measures:

a. Scale 1: Any article that failed the verification process as described above was classified as hallucination, assigning a score of 1 (hallucination) or 0 (real) to each reference.

$$s^{(1)} = \begin{cases} 1 & \text{if the citation hallucinated} \\ 0 & \text{if the citation is real} \end{cases}$$

b. Scale 2: Articles were also categorized into four categories based on the degree of hallucination: 25%, 50%, 75%, and 100%, depending on how many of the four key elements (article title, journal name, authors, publication year) were determined to be inaccurate.

$$s^{(2)} = \begin{cases} \frac{m}{4} & \text{if m elements are hallucinated (m=0,..,4)} \end{cases}$$

where *m=0* corresponds to the real citation.

We employed a decision tree, as depicted in Figure 1, to evaluate the accuracy of each of the *m* elements within a citation: [Title, Author(s), Journal, Date]. The initial step involved checking the title. If the title is verified as existent, it becomes the reference point, or "stem," for subsequent checks against the Author(s), Journal, and Date. In this scenario, any configuration beginning with [0,...] is possible, where '0' indicates the presence of a real title and subsequent zeros or ones represent the accuracy of the other elements in varying combinations. If the title is nonexistent, the tree advances to check the author(s). Should the Author(s) be verified, Author(s) then serves as the stem, and the tree proceeds to assess the Journal and Date. Any positive result here starts with a '1' (for the hallucinated title) followed by a '0' (for real Author(s)), and so forth. If neither the title nor the author(s) are confirmed, the citation is considered invalid, resulting in a configuration of [1,1,1,1] (all hallucinated), signifying the absence of all core components (Node 5 in Figure 1). This decision tree ensures a systematic and sequential assessment of the citation's validity.

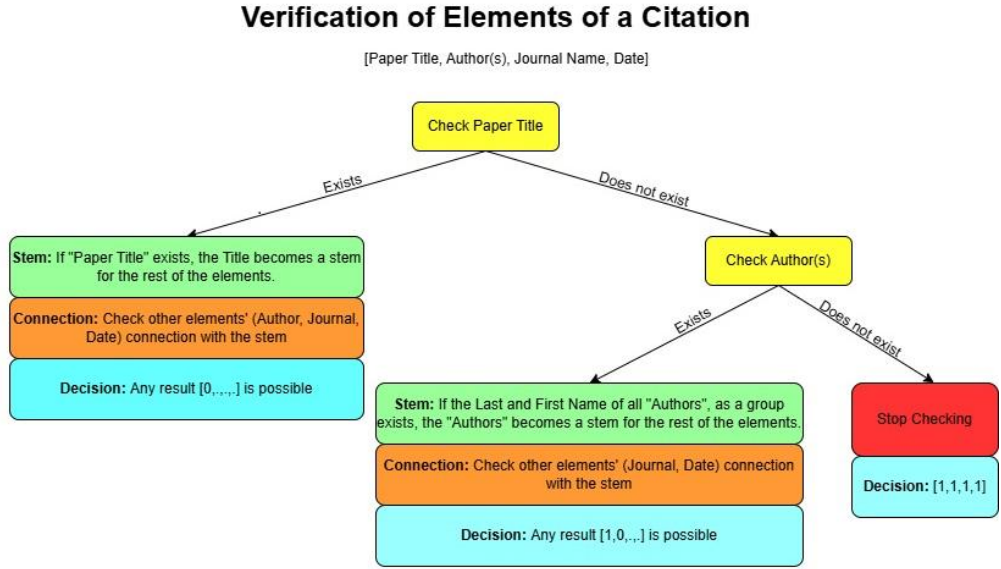

**Figure 1:** Decision Tree for Citation Validation. This diagram illustrates the sequential process of validating the elements of a citation—title, author(s), journal, and date—using a decision tree approach. Each node represents a decision point that determines the next step in the validation process based on the existence or absence of key citation components.

To ensure a reliable error rate, we selected 15 topics and instructed each chatbot to generate 10 references per topic. This process resulted in a total of 150 references for each chatbot. We assessed the hallucination rates by averaging these rates across the 150 citations for each of the three measurement scales. Consequently, we calculated two hallucination rates for each chatbot as follows:

$$\text{For Scale 1: } H^{(1)} = \sum_{i=1}^{15} \sum_{j=1}^{10} s_{ij}^{(1)} / 150$$

$$\text{For Scale 2: } H^{(2)} = \sum_{i=1}^{15} \sum_{j=1}^{10} s_{ij}^{(2)} / 150$$

where $i=1,\ldots,15$ denotes the topics, and $j=1,\ldots 10$ represents the ten references provided for each topic. By construction $H^{(2)} \leq H^{(1)}$, for all chatbots.

### 3.2 Assessing the Effect of Topic Recency on Hallucination

To test the relationship between topic recency and hallucination rates, we created a recency metric, $Recency_i$, and categorized the topics into "old" and "new" based on this metric. We then investigated the hallucination rates across these categories.

First, we compute the average age of any topic $i$ as follows:

$$Age\ (chatbot\ k)_i = \frac{\sum_{j=1}^{n} Age(chatbot\ k)_{i,j}}{n}$$

where $k=1, 2,$ and $3$ refer to each of the three chatbots (ChatGPT-4o, o1-preview, and Gemini A., respectively). $Age(chatbot\ k)_{i,j}$ represents the age of the real paper $j$ cited in topic $i$, and n is the total number of real references for topic $i$. For instance, if ChatGPT-4o cites three real papers from 2018, 2019, and 2020 for topic 2, then considering the current year is 2024, the age of that specific topic is $Age(ChatGPT)_{Topic\ 2} = \frac{6+5+4}{3} = 5$. If all 10 citations for a particular topic $i$ are hallucinated (which is the case with Gemini A. citations for topics 3, 6, 7, 8, 9, 11, and 15), then $Age(Gemini)_i$ for those references is undefined.

Thus, to calculate the recency of a topic, we rely solely on real citations. The average age for topic $i$, incorporating all chatbots, is given as follows:

$$Average\ age_i = \frac{\sum_{k=1}^{m} Age(chatbot\ k)_i}{m}\ for\ all\ well\ defined\ m\ chatbots\ with\ m \leq 3$$

For example, for topic 3, all of the citations provided by Gemini A. are hallucinated; thus, $Age(Gemini)_3$ is undefined. Therefore, we have $m=2$ well-defined chatbots. Thus, $Average\ age_3 = \frac{\sum_{k=1}^{2} Age(chatbot\ k)_3}{2}$.

In our experimental setting, $Age(ChatGPT)_i$ has been well defined since there is at least one real (non-hallucinated) citation for all $i$. As previously mentioned, this is not the case for Gemini A. Therefore, when $Age(Gemini)_i$ is undefined for topic $i$, $Average\ age_i$ is calculated solely using the well-defined ages from other chatbots.

Finally, the recency of any topic is assessed based on whether $Average\ age_i$ is less than 10 years. Topics with an average age of less than 10 years are considered '*New*', while those with an average age of 10 years or more are defined as '*Old*':

$$Recency_i = \begin{cases} New & if\ Average\ age_i < 10 \\ Old & Otherwise \end{cases}$$

This recency measure helps us classify whether our 15 topics are new.

## 4. Results

### Hallucination Rates

We first start by presenting the average hallucination rates of the three chatbots—ChatGPT-4o, ChatGPT-o1-preview, and Gemini A.—across 15 instances and two different scales, as shown in Figure **2**.

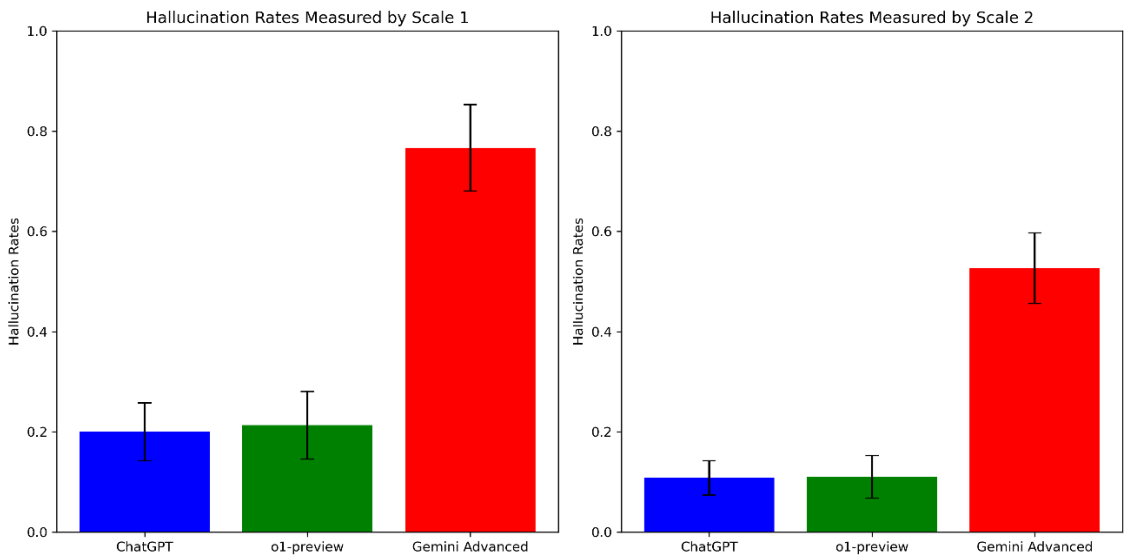

**Figure** 2: **Hallucination rates across chatbots on Scale 1 and Scale 2:** The charts compare the average hallucination rates of ChatGPT-4o, o1-preview, and Gemini Advanced across two different scales. The black lines represent the standard error of the mean, illustrating the variability in hallucination rates observed within each method, incorporating standard errors. On both scales, the hallucination rates for ChatGPT-4o and o1-preview were closely aligned and were consistently lower than those observed for Gemini Advanced. This trend underscores the significantly higher tendency for hallucinations to occur in Gemini Advanced patients than in ChatGPT-4o patients across both measurement scales.

By analyzing the response rates of the chatbots, we found that ChatGPT-4o had a response rate of 20.0% on Scale 1 (95% CI, 13.6%-26.4%). The hallucination rate for the o1-preview method was slightly higher at 21.3% (95% CI, 14.8%-27.9%). In contrast, Gemini A. exhibited a significantly greater percentage (76.7%) on the same scale (95% CI, 69.9%-83.4%).

Additionally, on Scale 2, ChatGPT-4o exhibited a hallucination rate of 10.8% (95% CI, 4.6%-13.9%), while o1-preview displays 11.0% (95% CI, 6.0%-16.0%), and Gemini A. had a hallucination rate of 52.7% (95% CI, 44.7%-60.7%). The lower rates under Scale 2, $H^{(2)}$, are due to the scale's allowance for partial hallucinations, in contrast to the rates under Scale 1, $H^{(1)}$, which categorizes any hallucination as 100%.

These results, together with the confidence intervals, reveal two main results. First, we cannot reject the hypothesis that ChatGPT-4o and o1-preview have the same hallucination rate. Second, Gemini A. has a higher hallucination rate than both versions of the ChatGPT.

Recency

We further wanted to test whether the incidence of hallucination varied with the recency of the field. The recency of topics is given in the "Recency" column of Table 2. For instance, in reference to "Topic 8: Cybersecurity in Financial Services," which is classified as a "New" topic, the highest hallucination rates for chatbots were 80%, 80%, and 100% on Scale 1 for ChatGPT-4o, o1-preview and Gemini A., respectively. In contrast, "Topic 1: Financial incentives," identified as an "Old" topic, had the lowest hallucination rates, with no hallucinations reported on either scale for all chatbots. To explore this further, we formulate a set of hypotheses, with the corresponding test results given in Table 1.

H.1.a: $H^{(1)}(ChatGPT-4o, Recency = New) = H^{(1)}(ChatGPT-4o, Recency = Old)$

H.1.b: $H^{(2)}(ChatGPT-4o, Recency = New) = H^{(2)}(ChatGPT-4o, Recency = Old)$

The second set of hypotheses belongs to the o1-preview:

H.2.a: $H^{(1)}(o1-preview, Recency = New) = H^{(1)}(o1-preview, Recency = Old)$

H.2.b: $H^{(2)}(o1-preview, Recency = New) = H^{(2)}(o1-preview, Recency = Old)$

The last set belongs to Gemini A:

H.3.a: $H^{(1)}(Gemini\ A., Recency = New) = H^{(1)}(Gemini\ A., Recency = Old)$

H.3.b: $H^{(2)}(Gemini\ A., Recency = New) = H^{(2)}(Gemini\ A., Recency = Old)$

|  | ChatGPT 4o | | o1-preview | | Gemini Advanced | |
|---|---|---|---|---|---|---|
|  | Scale 1 | Scale 2 | Scale 1 | Scale 2 | Scale 1 | Scale 2 |
| Old Topics | 11.3% | 4.7% | 7.5% | 2.8% | 72.5% | 41.9% |
| New Topics | 30.0% | 17.9% | 37.1% | 20.4% | 81.4% | 65.0% |
| Test of equality (t-stat) | 1.64 | 2.03 | 2.45 | 6.00 | 0.51 | 1.74 |
| Test of equality (p value) | 0.14 | 0.08 | 0.04 | 0.07 | 0.62 | 0.11 |

**Table 1: The effect of Recency of a Topic on Hallucination:** This table shows the hallucination rates for new and old topics across the chatbots under Scale 1 and Scale 2. Six tests of equality evaluate whether hallucination rates are identical between new and old topics for each chatbot across the scales. Even though the results do not indicate a clear pattern, one test has a p value of less than 5%, and three tests have a p value of less than 10%. The p value of the three tests is higher than 10%.

The first two hypotheses (H.1.a and H.1.b) assert that the recency of the topic (being new or old) has no impact on the hallucination rates of ChatGPT-4o, evaluated across two different metrics ($H^{(1)}$ and $H^{(2)}$). Similarly, the third and fourth hypotheses (H.2.a and H.2.b) claim that the recency of the topic has no impact on the hallucination rates of o1-preview-4o, evaluated across two different metrics. Finally, the fourth and fifth hypotheses (H.3.a and H.3.b, respectively) posited that the recency of the topic does not affect the hallucination rates of Gemini A., which were also assessed across two metrics. Even though hallucination rates are higher for newer topics than for older topics across all scales and chatbots, as shown in Table 1, statistical tests reveal nuanced differences. The results of t tests performed using the ChatGPT-4o and o1-preview models differed clearly from those conducted with the Gemini A. model. According to the ChatGPT-4o and o1-preview analyses, the findings are statistically significant at the 10% level in three out of four cases, with p values of 0.14 and 0.08, respectively. However, the significance decreases at the 5% level, with only one case remaining significant.

Specifically, while the hallucination rates for ChatGPT-4o are higher in newer topics, the p values are 0.14 and 0.08 for scales 1 and 2, respectively. Consequently, hypothesis H.1.a is not rejected at the 10% significance level, while H.1.b is. Neither hypothesis is rejected at the 5% level, indicating an ambiguous difference in hallucination rates between older and newer topics.

T tests conducted with o1-preview yield clearer, albeit not strong, results: While hallucination rates for newer topics are significantly greater than those for older topics under scale 1 at the 5% significance level (p value = 0.04), they do not reach statistical significance under scale 2 (p value = 0.07). Thus, both hypotheses are rejected at the 10% level; however, hypothesis H.2.a is not rejected, while H.2.b is rejected at the 5% level.

The t-tests conducted using the Gemini A. model produced definitive results. Hypotheses H.3.a and H.3.b cannot be rejected, with p values of 0.62 and 0.11, respectively. Thus, we can safely say that there is no difference in Gemini A.'s hallucination rates between new and old topics. This might be due to the consistently high hallucination rates of Gemini A., which obscures any discernible effects of topic recency.

Table 2 lists all the topics together with their corresponding hallucination rates and recency. The last three rows summarize the average hallucination rates together with the upper and lower bounds.

| | Chatbot Model | | ChatGPT 4o | | o1-preview | | Gemini Advanced | |
|---|---|---|---|---|---|---|---|---|
| Id | Topic | Recency | Scale 1 | Scale 2 | Scale 1 | Scale 2 | Scale 1 | Scale 2 |
| 1 | Financial Incentives | Old | 0.00 | 0.00 | 0.00 | 0.00 | 0.00 | 0.00 |
| 2 | The Regulation of AI in Finance | New | 0.50 | 0.33 | 0.40 | 0.20 | 0.80 | 0.60 |
| 3 | Cryptocurrency Regulation and Adoption | New | 0.10 | 0.08 | 0.20 | 0.05 | 1.00 | 0.75 |
| 4 | ESG Investing | New | 0.10 | 0.03 | 0.20 | 0.10 | 0.10 | 0.08 |
| 5 | Fintech Innovation and Disruption | New | 0.20 | 0.13 | 0.10 | 0.05 | 0.90 | 0.68 |
| 6 | Impact of Inflation on Investment | Old | 0.20 | 0.10 | 0.10 | 0.03 | 1.00 | 0.65 |
| 7 | Quantitative Easing and Its Aftermath | Old | 0.20 | 0.08 | 0.20 | 0.08 | 1.00 | 0.48 |
| 8 | Cybersecurity in Fin. Services | New | 0.80 | 0.43 | 0.80 | 0.40 | 1.00 | 0.98 |
| 9 | Decentralized Finance (DeFi) | New | 0.40 | 0.28 | 0.80 | 0.58 | 1.00 | 0.78 |
| 10 | Global Debt Crisis | Old | 0.30 | 0.15 | 0.10 | 0.05 | 0.60 | 0.33 |
| 11 | Credit Scoring | Old | 0.10 | 0.03 | 0.00 | 0.00 | 1.00 | 0.65 |
| 12 | Behavioral Finance | Old | 0.00 | 0.00 | 0.00 | 0.00 | 0.50 | 0.25 |
| 13 | Disruption in Investment Banking | New | 0.00 | 0.00 | 0.10 | 0.05 | 0.90 | 0.70 |
| 14 | Optimal investment and consumption | Old | 0.10 | 0.03 | 0.00 | 0.00 | 0.70 | 0.38 |
| 15 | Regulation after 2008 fin. crisis | Old | 0.00 | 0.00 | 0.20 | 0.08 | 1.00 | 0.63 |
| | Hallucination Rate | Average | 20.0% | 10.8% | 21.3% | 11.0% | 76.7% | 52.7% |
| | | Upper B. | 26.4% | 15.8% | 27.9% | 16.0% | 83.4% | 60.7% |
| | | Lower B. | 13.6% | 5.9% | 14.8% | 6.0% | 69.9% | 44.7% |

**Table 2:** Hallucination rates of ChapGPT-4o, o1-preview, and Gemini Advanced by topic under two different scales, binary (scale 1) and nonbinary (scale 2).

Overall, OpenAI's chatbots, ChatGPT-4o and o1-preview, tend to exhibit higher rates of hallucination for newer topics. In contrast, the consistently high hallucination rates observed with the Gemini A. model mask any patterns related to the recency of topics.

## 5. Conclusion and Discussion

Many researchers have documented that chatbots hallucinate. In this study, we aimed to perform several tasks. First, we compared the hallucination rate of a newly released chatbot, o1-preview, with that of ChatGPT-4o and Gemni Advanced. Second, we propose several new metrics for hallucination studies: one is a nonbinary metric, and the other is a recency index. We test the performance of these metrics with the aforementioned three chatbots. For the first task, we asked ChatGPT-4o, o1-preview, and Gemini Advanced to provide ten unique references for fifteen topics in finance. By analyzing 150 citations given by each chatbot and using two different metrics, our findings yielded several results. First and foremost, we documented the hallucination rates of three chatbots on binary and nonbinary scales. The hallucination rate of ChatGPT-4o under the binary metric (hallucination or no-hallucination) was 20.0% (95% CI, 13.6%-26.4%), and that under the binary metric was 10.8% (95% CI, 4.6%-13.9%). The hallucination rate of the o1-preview under the binary metric was 21.3% (95% CI, 14.8%-27.9%), and that under the binary metric was 11.0% (95% CI, 6.0%-16.0%). Thus, the o1-preview, which was released by the same company, Open AI, for the purpose of improving ChatGPT-4o does not yield better results than ChatGPT-4o. In regard to Gemini Advanced, which was developed by Google, the results are even worse. The hallucination rate for Gemini Advanced under the binary metric was 76.7% (95% CI, 69.9%-83.4%). With respect to the nonbinary metric, it decreased to 52.7% (95% CI, 44.7%-60.7%). Taken together, these findings provide our first broader result: Hallucinations occur significantly more frequently with Gemini Advanced than with either the ChatGPT-4o or o1- preview.

We further observed that the incidence of hallucination varied with the recency of the field. Even though hallucination rates are higher for newer topics than for older topics across all scales and chatbots, as shown in Table 1, statistical analyses reveal more subtle distinctions. The tests performed with OpenAI's ChatGPT-4o and o1-preview models are statistically significant at the 10% level in three out of four cases, but this significance diminishes at the 5% level, with only one out of four cases remaining significant. However, for Gemini Advanced, although a similar trend—higher hallucination rates in newer topics—is observed, it is statistically insignificant. This insignificance may primarily stem from the consistently high hallucination rates of Gemini Advanced, which may mask any potential effects of topic recency.

Researchers must exercise caution when using chatbots for citing financial literature, particularly when discussing newer topics. Gemini Advanced had significantly higher and more frequent hallucination rates than OpenAI ChatGPT-4o and o1-preview. This variability in accuracy underscores the importance of thorough verification of chatbot-provided references, especially in rapidly evolving fields.

## Declaration of competing interest

The authors declare that they have no known competing financial interests or personal relationships that could have appeared to influence the work reported in this paper.

## Declaration of generative AI and AI-assisted technologies in the writing process

During the preparation of this work, the author(s) used ChatGPT and Gemini Advanced to assist with drafting the initial content and language refinement. After using this tool/service, the author(s) reviewed and edited the content as needed and take(s) full responsibility for the content of the publication.